%% file: iclr_rlgm_soumission.tex
\documentclass{article} % For LaTeX2e
\usepackage{iclr2019_conference}
\usepackage{times}

\input{math_commands.tex}

\usepackage{hyperref}
\usepackage{url}

\title{Variational Recurrent Neural Networks for Graph Classification}

\author{Edouard Pineau$^{\star \dag}$ \& Nathan de Lara$^{\star}$ \\
Telecom ParisTech$^{\star}$, Safran$^{\dag}$\\
\texttt{\{edouard.pineau,nathan.delara\}@telecom-paristech.fr} \\
}

\renewcommand{\bibname}{References}
\renewcommand{\bibsection}{\subsubsection*{\bibname}}

\usepackage[english]{babel}
\addto{\captionsenglish}{%
  \renewcommand{\bibname}{References}
}
\usepackage[utf8x]{inputenc}
\usepackage[T1]{fontenc}
\usepackage[noend]{algpseudocode}
\usepackage{fancyhdr,amssymb}
\usepackage[ruled,vlined]{algorithm2e}
\usepackage{array}
\usepackage{stmaryrd}
\usepackage{amsfonts}

\usepackage{amsfonts,amsmath,amssymb,amsthm,epsfig,psfrag}
\usepackage[T1]{fontenc}
\usepackage{babel}

\usepackage{amsmath}
\usepackage{mathtools}
\usepackage{dsfont}
\usepackage{systeme}
\usepackage{graphicx}
\usepackage[colorinlistoftodos]{todonotes}
\usepackage{blindtext}
\usepackage{enumitem}
\usepackage{xcolor}

\newcommand{\cmmnt}[1]{}

\newcolumntype{M}[1]{>{\centering\arraybackslash}m{#1}}
\newcolumntype{N}{@{}m{0pt}@{}}

\setlist[description]{leftmargin=\parindent,labelindent=\parindent}

\begin{document}

\maketitle

\begin{abstract}
    We address the problem of graph classification based only on structural information. Inspired by natural language processing techniques (NLP), our model sequentially embeds information to estimate class membership probabilities. Besides, we experiment with NLP-like variational regularization techniques, making the model predict the next node in the sequence as it reads it. We experimentally show that our model achieves state-of-the-art classification results on several standard molecular datasets. Finally, we perform a qualitative analysis and give some insights on whether the node prediction helps the model better classify graphs.
    
\end{abstract}

\section{Introduction}

Many natural or synthetic systems have a natural graph representation where entities are described through their mutual connections: chemical compounds, social or biological networks, for example. Therefore, automatic mining of such structures is useful in a variety of applications.

%Graphs can be studied either individually, considering the nodes as samples, or collectively, each sample of the dataset being a graph object. Here, we consider the later case, applied to classification task. This setting
Graph classification raises several difficulties to leverage standard machine learning algorithms. Indeed, most of these algorithms take vectors of fixed size as inputs. In the case of graphs, usual representations such as edge list or adjacency matrix do not match this constraint. The size of the representations is graph dependent (number of edges in the first case, number of nodes squared in the second) and these representations are index dependent: up to indexing of its nodes, a same graph admits several equivalent representations. In a classification task, the label of a graph is independent from the indices of its nodes, so the model used for prediction should be invariant to node ordering as well. Handling discrete inputs with variable size and ordering is a well known problem in natural language processing (NLP). This is why we adapt NLP techniques to tackle graph classification.

In this paper, we propose a method to sequentially embed graph information in order to perform classification. By construction, this recurrent graph classifier overcomes the common difficulties listed above. Besides, we propose to use an additional node prediction block to help the model to capture the intrinsic structure of the graphs. The complete model is denoted \textit{variational recurrent graph classifier} (VRGC). Experiments show that this leads to better classification results for larger datasets. For clarity of the contribution of our work, we use neither node attributes nor edge attributes.

Related works use either graph kernels \citep{nikolentzos2017kernel, nikolentzos2017matching, nikolentzos2018degeneracy, neumann2016propagation, shervashidze2011weisfeiler, yanardag2015deep}, sequential methods \citep{callut2008classification, xu2012protein, jin2018learning, you2018graphrnn} or graph features \citep{barnett2016feature, DBLP:journals/corr/NarayananCVCLJ17, gomez2017dynamics, dutta2017high}.

\section{Model}
\label{sec:model}

We propose to use a sequential approach to embed graphs with a variable number of nodes and edges into a vector space of a chosen dimension. This latent representation is then used for classification. Node index invariance is approximated through specific pre-processing and aggregation.

Let $G=(V, E)$ be an undirected and unweighted graph with $V$ a set of nodes and $E$ a set of edges. The graph $G$ can be represented, modulo any permutation $\pi$ over its nodes $\{n_i\}_{i=1}^{|V|}$, by its boolean adjacency matrix $A^{\pi}$ such that $A^{\pi}_{ij} = 1$ if nodes indexed by $i$ and $j$ are connected in the graph and $A^{\pi}_{ij} = 0$ otherwise. We use this adjacency matrix as a raw representation of the graph. 

\begin{figure*}
    \centering
    \includegraphics[width=1\textwidth]{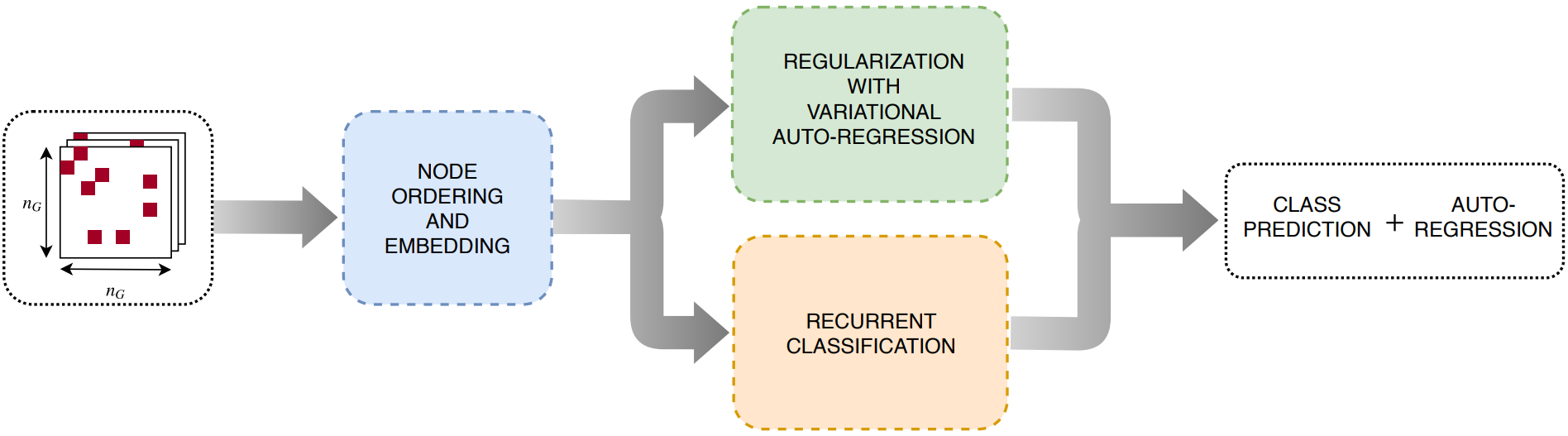}
    \caption{Macroscopic representation of VRGC.}
    \label{fig:macro}
\end{figure*}

Our VRGC is composed of three main parts: node ordering and embedding, classification and regularization with variational auto-regression (VAR). See Figure \ref{fig:macro} for an illustration. % Each of these parts is detailed respectively in the following paragraphs.

\paragraph{Node ordering and embedding}
\label{subsec:node ordering}

Before being processed by the neural network, the adjacency matrix of a graph is transformed on-the-fly \citep{you2018graphrnn}. First, a node is selected at random and used as root for a breadth first search (BFS) over the graph. The rows and columns of the adjacency matrix are then reordered according to the sequence of nodes returned by the BFS. Next, each row $i$ (corresponding to the $i^{th}$ node in the BFS ordering) is truncated to keep only the connections of node $n_{i}$ with the $\min(i, d)$ nodes that preceded in the BFS. This way, each node is $d$-dimensional, and each truncated matrix is zero-padded in order to have dimensions $(|V|_{max}, d_n)$. Throughout the rest of the paper, we use the notation $n_G$ for $|V|_{max}$. 

After node ordering and pre-embedding, each graph is processed as a sequence of $d$-dimensional nodes by a gated recurrent unit (GRU) neural network \citep{cho2014learning}. The GRU is a special RNN able to learn long term dependencies by solving vanishing gradient effect\footnote{The choice of GRU over Long Short Term Memory networks is arbitrary as they have equivalent long-term modeling power \citep{chung2014empirical}.}. In order to help the recurrent network training, we propose to add a simple fully connected network between pre-embedding and recurrent embedding. Therefore, the node will be presented to the GRU in the shape of continuous vectors instead of binary adjacency vectors. 

\begin{figure*}
    \centering
    \includegraphics[width=1\textwidth]{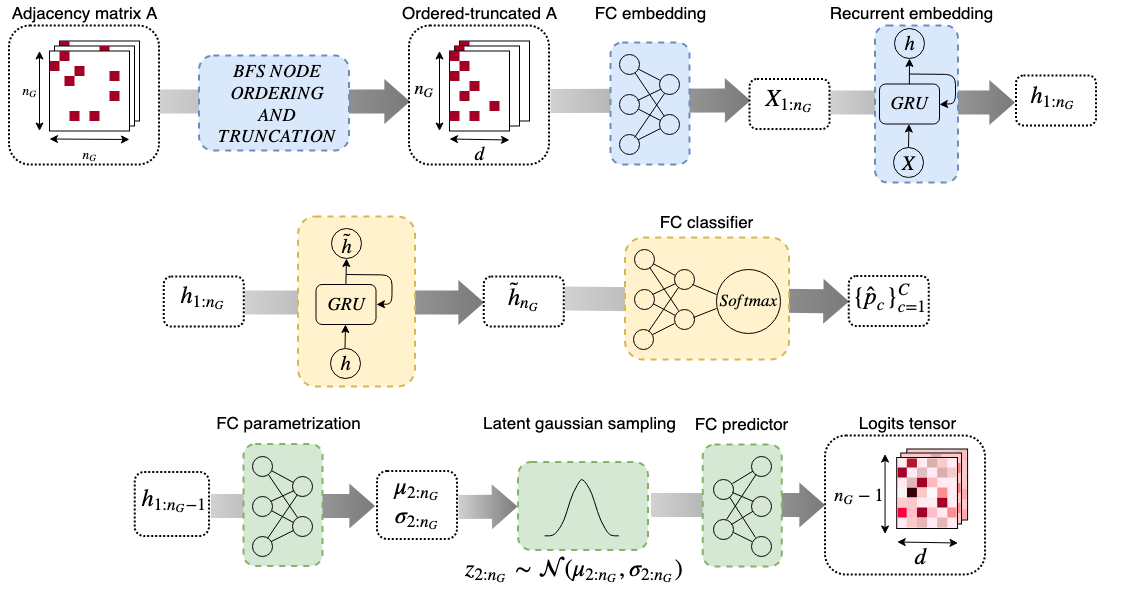}
    \caption{Architecture for VRGC. Top: node ordering and embedding. Middle: classification. Bottom: regularization with VAR plus final aggregation. FC stands for fully-connected.}
    \label{fig:archi}
\end{figure*}

Finally the GRU sequentially embeds each node $n_i$ by using $n_{i-1}$ and information contained in a memory cell $h_{i-1}$ that theoretically embeds all previously seen information. The embedded node sequence $\{h_i\}_{i=1}^{n_G}$ then feeds both the VAR and the classifier as discussed in subsequent sections. See top line of Figure \ref{fig:archi}.

\paragraph{Classification}
\label{subsec:classification}

After the embedding step, we use an additional GRU dedicated to classification that takes $\{h_i\}_{i=1}^{n_G}$ as input.
Its last memory cell, denoted $\tilde{h}_{n_G}$, feeds a softmax multilayer perceptron (MLP) which performs class prediction.
Formally, let $c$ be the class index, the classifier is trained by minimizing the cross-entropy loss between ground-truth and $\hat{p}(G, r)$ the softmax class membership probability vector for a given graph $G$ that has been sorted by a BFS rooted with node $r$. We call this objective term $\mathcal{L}_{classif}$.
As discriminating patterns might be spread across the whole graph, the network is required to model long-term dependencies. By construction, GRUs have such ability. See middle line of Figure \ref{fig:archi}.

\paragraph{Regularization with variational auto-regression}
\label{subsec:vap}

As the structure of a graph is the concatenation of the interactions between all nodes and their respective neighbors, learning a good representation without using node attributes requires for the model to capture the structure of the graph while classifying. Accordingly, we add an auto-regression block to our model: at each node, the network makes a prediction for the next node adjacency. Multi-task learning is a powerful leverage to learn rich representation \citep{sanh2018hierarchical} in NLP. In particular, such representation for sequence classification has already been used for sentiment analysis \citep{latif2017variational, xu2017variational}. %It is the equivalent of predicting the $i^{th}$ word of a sentence, given an aggregated representation of this sentence up to word $i-1$. 

We use a variational auto-encoder (VAE) \citep{kingma2013auto} to learn a representation of each node $n_i$ given $h_{i-1}$. The first layers of the encoder are shared with the classifier and corresponds to the graphs preprocessing (blue part in Figures \ref{fig:macro} and \ref{fig:archi}). The subsequent encoder layers, the latent sampling and the decoder constitute the VAR. For each graph $G$ with embedded nodes $\{ n_{i} \}_{i \in \llbracket 1, n_G \rrbracket}$, the fully connected variational auto-encoder takes $h\coloneqq\{ h_{i} \}_{i \in \llbracket 1, n_G-1 \rrbracket }$ as input. Let $\{ z_{i} \}_{i \in \llbracket 1, n_G \rrbracket}$ be the latent random variables for the following model 

\begin{align*}
    p(G,z|h) = p(n_1, z_1)\prod_{i=2}^{n_G}{p_\theta(n_i|z_{i})q_\phi(z_{i}|h_{i-1})}.
\end{align*} 

In practice, $p_\theta$ and $q_\phi$ are modelled by neural networks parametrized by $\theta$ and $\phi$, which require differentiable functions for training. However, $p_\theta(n_i|z_i)$ models a binary adjacency vector representing the connections between node $n_i$ and previously visited nodes $n_{j<i}$. Therefore, we use sigmoid continuous relaxation to train our model, and hard binary sampling at test time. We use a Gaussian variational posterior distribution. Training is done by maximizing the variational lower bound of the log-likelihood of the observation as in Kingma's VAE. The exact loss is displayed in Appendix \ref{app:elbo} and denoted $\mathcal{L}_{pred}$. 

The regularization part is illustrated in the bottom line of Figure \ref{fig:archi}. In the end, the model is trained by minimizing the total loss $\mathcal{L} = \mathcal{L}_{classif} + \alpha \mathcal{L}_{pred}$, where $\alpha$ is a hyper-parameter.

\paragraph{Aggregation of the results at test time}
\label{sub:aggreg}

The node ordering step introduces randomness to our model. On the one hand, it helps learn more general graph representations during the training phase, but on the other hand, it might produce different outputs for the same graph during the testing phase, depending on the root of the BFS. In order to counter this side effect, we add the following aggregation step for the testing phase\cmmnt{ combining a soft and a hard vote}. Each graph is processed $N$ times by the model with $N$ different roots for BFS ordering. The $N$ class membership probability vectors are extracted and averaged. The average score vector is noted $\bar{p}$ and computed as follows with an element-wise sum: 

%\vspace{-0.5cm}
%\begin{align*}
%\label{eq:average}
$$
    \bar{p}(G) = \frac{1}{N} \sum_{\substack{i=1 \\ r \sim \mathcal{U}(\llbracket 1, n_G \rrbracket)}}^N {\hat{p}(G, r)}.
$$
%\end{align*}
%\vspace{-0.5cm}

This soft vote is repeated $K$ times resulting in $K$ probability vectors $\left\{\bar{p}_{.,k}(G)\right\}_{k=1}^K$ for each graph $G$. The final class attributed to a graph corresponds to the highest probability among the $K$ vectors.  
This second hard vote enables to choose the batch of votes for which the model is the most confident.
%\vspace{-0.4cm}
\begin{align*}
    \hat{c}(G) & = \arg \max_{c\in \llbracket 1,C \rrbracket} \left\{ || \bar{p}_{c, .}(G) ||_\infty \right\}
\end{align*}

\section{Experiments}
\label{sec:experiments}

\paragraph{Datasets and results}

We evaluated our model against four standard datasets from biology: Mutag (MT), Enzymes (EZ), Proteins Full (PF) and National Cancer Institute (NCI1) \citep{KKMMN2016}. A detailed description of each dataset in provided in Appendix \ref{app:datasets}.

We compare our results to those obtained by Earth Mover's Distance \citep{nikolentzos2017matching} (EMD), Pyramid Match \citep{nikolentzos2017matching} (PM), Feature-Based \citep{barnett2016feature} (FB), Dynamic-Based Features \citep{gomez2017dynamics} (DyF), Stochastic Graphlet Embedding \citep{dutta2017high} (SGE), Truncated Laplacian Spectrum \citep{de2018simple} (TLS) and Family of Graph Spectral Distances \citep{verma2017hunt} (FGSD). All values are directly taken from the aforementioned papers as they use a setup similar to ours. For algorithms presenting results with and without node features, we reported the results without node features. For those presenting results with several sets of hyper-parameters, we reported the results for the parameters that performed best on the largest number of datasets. Results are reported in Table \ref{tab:results}. We obtain state-of-the-art results on three out of the four datasets used for this paper and the second best result on the fourth one. 

\begin{table}
    \centering
    \renewcommand{\arraystretch}{1.2}
    \begin{tabular}{l|M{1.5cm} M{1.5cm} M{1.5cm} M{1.5cm} M{1.5cm} M{1.5cm}}
           & MT            &      EZ       & PF            & NCI1 \\
      \hline
      EMD  & 86.1          &     36.8      & -             & 72.7 \\
      PM   & 85.6          &     28.2      & -             & 69.7 \\
      FB   & 84.7          &     29.0      & 70.0          & 62.9 \\
      DyF  & 86.3          &     26.6      & 73.1          & 66.6 \\
      SGE  & 87.3          &     40.7      & 71.9          & - \\
      TLS  & 88.4          &     43.7      & 73.6          & 75.2 \\
      FGSD & \textbf{92.1} &     -         & 73.4          & 79.8 \\
      \hline
      RGC  & 89.5          & \textbf{48.7} & 72.5          & 78.1 \\
      VRGC & 86.3          & 48.4          & \textbf{74.8} & \textbf{80.7} \\
      \end{tabular}
  \caption{Experimental results of different models plus our own on four standard molecular datasets. RGC stands for recurrent graph classifier, VRGC for variational RGC. All other acronyms are defined in section \ref{sec:experiments}.}
  \label{tab:results}
\end{table}

\paragraph{Node indexing invariance}
Our model is designed to be independent from node ordering of the graph with respect to different BFS roots. Inputs representing the same graph (up to node ordering) should be close from one another in the latent embedding space. As the preprocessing is performed on each graph at each epoch, a same graph is processed many times by the model during training with different embeddings. This creates a natural regularization for the network. Indeed, as illustrated in Figure \ref{fig:projection}, the projections corresponding to the same graphs form a heap in the low dimensional representation of the latent space.

\paragraph{Contribution of the VAR to classification}
The variational regularization term seems to help the model finding a more meaningful latent representation for classification while graph dataset becomes larger. Note that the extra cost of training the VAR is marginal with respect to the training of the RNNs. We provide an illustration of the output of VAR block in Figure \ref{fig:reconstructions}.

\paragraph{Conclusion and room for improvement}
This paper proposed a recurrent graph classifier with variational regularization. The invariance to node indexing is greedy learned from numerous iterations on randomly rooted BFS-ordered graph. For future work, we should investigate the impact VAR capacity (number and size of hidden layers) on classification accuracy or generalization.

\paragraph{Acknowledgments} We would like to thank Thomas Bonald and Sebastien Razakarivony for their comments and help. We also would like to thank NVIDIA and its GPU Grant Program for providing the hardware we used in our experiments. This work is supported by the company Safran through the CIFRE convention 2017/1317.

\bibliographystyle{abbrvnat}
\bibliography{biblio}

\clearpage
\appendix

\section{VAR loss}
\label{app:elbo}

The VAE-like loss for VAR regularization is the following:

\begin{align*}
    \mathcal{L}_{pred} = \mathbb{E}_{p_d(G)}\left[\sum_{i=2}^{n_G}{\mathrm{KL}\left( q_\phi(z_i|h_{i-1})||q(z_i)\right)}\right] 
    - \mathbb{E}_{p_d(G)}\left[\sum_{i=2}^{n_G}{\mathbb{E}_{q_\phi(z_{i}|h_{i-1})}\left[\log p_\theta(n_i|z_i)\right]}\right],
\end{align*}

which is a lower bound of the negative marginal log-likelihood $\mathbb{E}_{p_d(G)}\left[\log p_\theta(G)\right]$. $p_\theta$ and $q_\phi$ are the respective densities of $n|z$ and $z|h$, whose distribution are parameterized by $\theta$ and $\phi$ respectively. KL denotes the Kullback-Leibler divergence, $p_d$ is the empirical distribution of $G$ and $q(z_i)$ is the density of the prior distribution of latent variables $\{ z_i \}_{i=2}^{n_G}$. We chose the standard Gaussian prior for $q(z_i)$. 

\section{Dataset characteristics}
\label{app:datasets}
 All graphs represent chemical compounds, nodes are molecular substructures (typically atoms) and edges represent connections between these substructures (chemical bound or spatial proximity). In MT, the compounds are either mutagenic or not mutagenic. EZ contains tertiary structures of proteins from the 6 Enzyme Commission top level classes; it is the only multiclass dataset of this paper. PF is a subset of the Dobson and Doig dataset representing secondary structures of proteins being either enzyme or not enzyme. In NCI1, compounds either have an anti-cancer activity or do not. %Statistics about the graphs are presented in appendix \ref{app:datasets}. 

\begin{table}[h]
    \centering
    \renewcommand{\arraystretch}{1.4}
    \begin{tabular}{l|M{1.5cm} M{1.5cm} M{1.5cm} M{1.5cm} M{1.5cm} M{1.5cm}}
                          & MT   & EZ   & PF   &  NCI1 \\
        \hline
        $\#$ graphs       & 188  & 600  & 1113 & 4110 \\
        $\#$ classes      & 2    & 6    & 2    & 2    \\
        bias              & 0.66 & 0.17 & 0.60 & 0.5  \\
        avg. |V|          & 18   & 33   & 39   & 30 \\
        min |V| / max |V| & 10/28   & 2/125   & 4/620   & 3/106 \\
        avg. |E|          & 39   & 124  & 146  & 64.6 \\
    \end{tabular}
    \caption{Basic characteristics of the datasets. Bias indicates the proportion of the largest class.}
    \label{tab:datasets}
\end{table}

\section{Features of network architecture}
\label{app:architecture}

 MT, EZ, PF and NCI1 are respectively divided into 10 folds such that the class proportions are preserved in each fold for all datasets. These folds are then used for cross-validation i.e., one fold serves as the testing set while the other ones compose the training set. Results are averaged over all testing sets.
Our model is implemented in Pytorch \citep{paszke2017pytorch} and trained with the Adam stochastic optimization method \citep{kingma2014adam} on a NVIDIA TitanXp GPU. 

\begin{table}
\centering
  \begin{center}
  
    \begin{tabular}{l|l}
      \textbf{Step} & \textbf{Architecture}  \\
      \hline
       BFS          & 1-layer FC. $d_n \times 64$ \\
       embedding    & 2-layer GRU. $64 \times 128$ \\
       \hline
       VAR          & \underline{Encoder} \\
                    & 1-layer FC. $128 \times 2 \times 8$ \\
                    &       Gaussian sampling \\
                    & \underline{Predictor} \\ 
                    & 2-layer ReLU FC. $8 \times d_n$ \\
        \hline
       Classifier   & 2-layer GRU. $128 \times 128$ + DP(0.25) \\
                    & 2-layer ReLU FC. $128 \times C$ + SF \\
    \end{tabular}
  \end{center}
  \caption{Generic architecture used in our experiments. ReLU FC stands for fully-connected network with ReLU activation. DP stands for dropout. SF stands for softmax.}
  \label{tab:table1}
\end{table}

The input size $d_n$ of the recurrent neural network is chosen for each dataset according to the algorithm described in \citep{you2018graphrnn}, namely 11 for MT, 25 for EZ, 80 for PF and 11 for NCI1. $\alpha$ is set to $0.1$. For training, batch size is set to 64, and the learning rate to $10^{-3}$, decreased by $0.3$ at iterations $400$ and $1000$. We use the same hyper-parameters for every dataset.

\section{Illustration of experimental results}

The following table presents the hyperparametrization of our model, i.e. the neural network architectures for each part presented in Section \ref{sec:model}.

\begin{figure}[h]
    \centering
    \includegraphics[width=8cm]{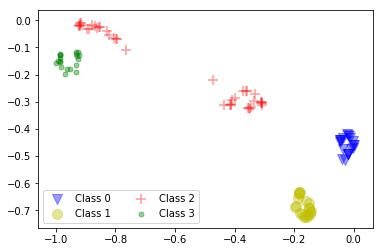}
    \caption{TSNE projection of the latent state preceding classification for five graphs of EZ each initiated with 20 different BFS. Colors and markers represent the respective classes of the graphs.}
    \label{fig:projection}
\end{figure}

\begin{figure}
    \centering
    \includegraphics[width=0.245\textwidth]{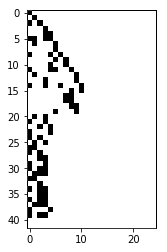}
    \includegraphics[width=0.245\textwidth]{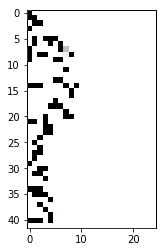}
    \includegraphics[width=0.245\textwidth]{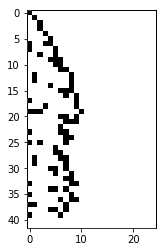}
    \includegraphics[width=0.245\textwidth]{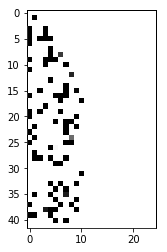}
    \caption{Left: Representation of the same graph after two differently rooted BFS ordering and truncation. Right: corresponding auto-regressions.}
    \label{fig:reconstructions}
\end{figure}

\end{document}

%% file: math_commands.tex
\usepackage{amsmath,amsfonts,bm}

\def\eqref#1{equation~\ref{#1}}
\def\1{\bm{1}}

\DeclareMathAlphabet{\mathsfit}{\encodingdefault}{\sfdefault}{m}{sl}
\SetMathAlphabet{\mathsfit}{bold}{\encodingdefault}{\sfdefault}{bx}{n}

%% file: iclr_rlgm_soumission.bbl
\begin{thebibliography}{25}
\providecommand{\natexlab}[1]{#1}
\providecommand{\url}[1]{\texttt{#1}}
\expandafter\ifx\csname urlstyle\endcsname\relax
  \providecommand{\doi}[1]{doi: #1}\else
  \providecommand{\doi}{doi: \begingroup \urlstyle{rm}\Url}\fi

\bibitem[Barnett et~al.(2016)Barnett, Malik, Kuijjer, Mucha, and
  Onnela]{barnett2016feature}
I.~Barnett, N.~Malik, M.~L. Kuijjer, P.~J. Mucha, and J.-P. Onnela.
\newblock Feature-based classification of networks.
\newblock \emph{arXiv preprint arXiv:1610.05868}, 2016.

\bibitem[Callut et~al.(2008)Callut, Fran{\c{c}}oisse, Saerens, and
  Dupont]{callut2008classification}
J.~Callut, K.~Fran{\c{c}}oisse, M.~Saerens, and P.~Dupont.
\newblock Classification in graphs using discriminative random walks, 2008.

\bibitem[Cho et~al.(2014)Cho, Van~Merri{\"e}nboer, Gulcehre, Bahdanau,
  Bougares, Schwenk, and Bengio]{cho2014learning}
K.~Cho, B.~Van~Merri{\"e}nboer, C.~Gulcehre, D.~Bahdanau, F.~Bougares,
  H.~Schwenk, and Y.~Bengio.
\newblock Learning phrase representations using rnn encoder-decoder for
  statistical machine translation.
\newblock \emph{arXiv preprint arXiv:1406.1078}, 2014.

\bibitem[Chung et~al.(2014)Chung, Gulcehre, Cho, and
  Bengio]{chung2014empirical}
J.~Chung, C.~Gulcehre, K.~Cho, and Y.~Bengio.
\newblock Empirical evaluation of gated recurrent neural networks on sequence
  modeling.
\newblock \emph{arXiv preprint arXiv:1412.3555}, 2014.

\bibitem[de~Lara and Pineau(2018)]{de2018simple}
N.~de~Lara and E.~Pineau.
\newblock A simple baseline algorithm for graph classification.
\newblock \emph{Relational Representation Learning Workshops (NIPS 2018)},
  2018.

\bibitem[Dutta and Sahbi(2017)]{dutta2017high}
A.~Dutta and H.~Sahbi.
\newblock High order stochastic graphlet embedding for graph-based pattern
  recognition.
\newblock \emph{arXiv preprint arXiv:1702.00156}, 2017.

\bibitem[Gomez et~al.(2017)Gomez, Chiem, and Delvenne]{gomez2017dynamics}
L.~G. Gomez, B.~Chiem, and J.-C. Delvenne.
\newblock Dynamics based features for graph classification.
\newblock \emph{arXiv preprint arXiv:1705.10817}, 2017.

\bibitem[Jin and JaJa(2018)]{jin2018learning}
Y.~Jin and J.~F. JaJa.
\newblock Learning graph-level representations with gated recurrent neural
  networks.
\newblock \emph{arXiv preprint arXiv:1805.07683}, 2018.

\bibitem[Kersting et~al.(2016)Kersting, Kriege, Morris, Mutzel, and
  Neumann]{KKMMN2016}
K.~Kersting, N.~M. Kriege, C.~Morris, P.~Mutzel, and M.~Neumann.
\newblock Benchmark data sets for graph kernels, 2016.
\newblock \url{http://graphkernels.cs.tu-dortmund.de}.

\bibitem[Kingma and Ba(2014)]{kingma2014adam}
D.~P. Kingma and J.~Ba.
\newblock Adam: A method for stochastic optimization.
\newblock \emph{arXiv preprint arXiv:1412.6980}, 2014.

\bibitem[Kingma and Welling(2013)]{kingma2013auto}
D.~P. Kingma and M.~Welling.
\newblock Auto-encoding variational bayes.
\newblock \emph{arXiv preprint arXiv:1312.6114}, 2013.

\bibitem[Latif et~al.(2017)Latif, Rana, Qadir, and Epps]{latif2017variational}
S.~Latif, R.~Rana, J.~Qadir, and J.~Epps.
\newblock Variational autoencoders for learning latent representations of
  speech emotion.
\newblock \emph{arXiv preprint arXiv:1712.08708}, 2017.

\bibitem[Narayanan et~al.(2017)Narayanan, Chandramohan, Venkatesan, Chen, Liu,
  and Jaiswal]{DBLP:journals/corr/NarayananCVCLJ17}
A.~Narayanan, M.~Chandramohan, R.~Venkatesan, L.~Chen, Y.~Liu, and S.~Jaiswal.
\newblock graph2vec: Learning distributed representations of graphs.
\newblock \emph{CoRR}, abs/1707.05005, 2017.
\newblock URL \url{http://arxiv.org/abs/1707.05005}.

\bibitem[Neumann et~al.(2016)Neumann, Garnett, Bauckhage, and
  Kersting]{neumann2016propagation}
M.~Neumann, R.~Garnett, C.~Bauckhage, and K.~Kersting.
\newblock Propagation kernels: efficient graph kernels from propagated
  information.
\newblock \emph{Machine Learning}, 102\penalty0 (2):\penalty0 209--245, 2016.

\bibitem[Nikolentzos et~al.(2017{\natexlab{a}})Nikolentzos, Meladianos, Tixier,
  Skianis, and Vazirgiannis]{nikolentzos2017kernel}
G.~Nikolentzos, P.~Meladianos, A.~J.-P. Tixier, K.~Skianis, and
  M.~Vazirgiannis.
\newblock Kernel graph convolutional neural networks.
\newblock \emph{arXiv preprint arXiv:1710.10689}, 2017{\natexlab{a}}.

\bibitem[Nikolentzos et~al.(2017{\natexlab{b}})Nikolentzos, Meladianos, and
  Vazirgiannis]{nikolentzos2017matching}
G.~Nikolentzos, P.~Meladianos, and M.~Vazirgiannis.
\newblock Matching node embeddings for graph similarity.
\newblock In \emph{AAAI}, pages 2429--2435, 2017{\natexlab{b}}.

\bibitem[Nikolentzos et~al.(2018)Nikolentzos, Meladianos, Limnios, and
  Vazirgiannis]{nikolentzos2018degeneracy}
G.~Nikolentzos, P.~Meladianos, S.~Limnios, and M.~Vazirgiannis.
\newblock A degeneracy framework for graph similarity.
\newblock In \emph{IJCAI}, pages 2595--2601, 2018.

\bibitem[Paszke et~al.(2017)Paszke, Gross, Chintala, and
  Chanan]{paszke2017pytorch}
A.~Paszke, S.~Gross, S.~Chintala, and G.~Chanan.
\newblock Pytorch, 2017.

\bibitem[Sanh et~al.(2018)Sanh, Wolf, and Ruder]{sanh2018hierarchical}
V.~Sanh, T.~Wolf, and S.~Ruder.
\newblock A hierarchical multi-task approach for learning embeddings from
  semantic tasks.
\newblock \emph{arXiv preprint arXiv:1811.06031}, 2018.

\bibitem[Shervashidze et~al.(2011)Shervashidze, Schweitzer, Leeuwen, Mehlhorn,
  and Borgwardt]{shervashidze2011weisfeiler}
N.~Shervashidze, P.~Schweitzer, E.~J.~v. Leeuwen, K.~Mehlhorn, and K.~M.
  Borgwardt.
\newblock Weisfeiler-lehman graph kernels.
\newblock \emph{Journal of Machine Learning Research}, 12\penalty0
  (Sep):\penalty0 2539--2561, 2011.

\bibitem[Verma and Zhang(2017)]{verma2017hunt}
S.~Verma and Z.-L. Zhang.
\newblock Hunt for the unique, stable, sparse and fast feature learning on
  graphs.
\newblock In \emph{Advances in Neural Information Processing Systems}, pages
  88--98, 2017.

\bibitem[Xu et~al.(2017)Xu, Sun, Deng, and Tan]{xu2017variational}
W.~Xu, H.~Sun, C.~Deng, and Y.~Tan.
\newblock Variational autoencoder for semi-supervised text classification.
\newblock In \emph{AAAI}, pages 3358--3364, 2017.

\bibitem[Xu et~al.(2012)Xu, Lu, He, Pan, and Jing]{xu2012protein}
X.~Xu, L.~Lu, P.~He, Z.~Pan, and C.~Jing.
\newblock Protein classification using random walk on graph.
\newblock In \emph{International Conference on Intelligent Computing}, pages
  180--184. Springer, 2012.

\bibitem[Yanardag and Vishwanathan(2015)]{yanardag2015deep}
P.~Yanardag and S.~Vishwanathan.
\newblock Deep graph kernels.
\newblock In \emph{Proceedings of the 21th ACM SIGKDD International Conference
  on Knowledge Discovery and Data Mining}, pages 1365--1374. ACM, 2015.

\bibitem[You et~al.(2018)You, Ying, Ren, Hamilton, and
  Leskovec]{you2018graphrnn}
J.~You, R.~Ying, X.~Ren, W.~L. Hamilton, and J.~Leskovec.
\newblock Graphrnn: A deep generative model for graphs.
\newblock \emph{arXiv preprint arXiv:1802.08773}, 2018.

\end{thebibliography}
